\documentclass[10pt,twocolumn,letterpaper]{article}

\usepackage{cvpr}
\usepackage{times}
\usepackage{epsfig}
\usepackage{graphicx}
\DeclareGraphicsExtensions{.pdf}
\usepackage{amsmath}
\usepackage{amssymb}
\usepackage{multirow}
\usepackage{url}
\usepackage[british,UKenglish,USenglish,english,american]{babel}

\usepackage{tabulary}
\usepackage{array}

\newcommand{\tms}{\cdot}
\newcommand{\str}{$^{'}$}
\newcommand{\nnum}{}

\newcommand{\ul}[1]{\underline{#1}}
\usepackage[bookmarks=false,colorlinks=true,linkcolor=black,citecolor=black,filecolor=black,urlcolor=black]{hyperref}

\cvprfinalcopy 

\def\cvprPaperID{***} 

\ifcvprfinal\fi
\begin{document}

\title{Convolutional Neural Networks at Constrained Time Cost}

\author{Kaiming He \qquad\qquad\qquad Jian Sun \vspace{8pt}\\
Microsoft Research\\
{\tt\small \{kahe,jiansun\}@microsoft.com}
}

\maketitle

\begin{abstract}
Though recent advanced convolutional neural networks (CNNs) have been improving the image recognition accuracy, the models are getting more complex and time-consuming. For real-world applications in industrial and commercial scenarios, engineers and developers are often faced with the requirement of constrained time budget. In this paper, we investigate the accuracy of CNNs under constrained time cost. Under this constraint, the designs of the network architectures should exhibit as trade-offs among the factors like depth, numbers of filters, filter sizes, \etc.
With a series of controlled comparisons, we progressively modify a baseline model while preserving its time complexity. This is also helpful for understanding the importance of the factors in network designs. We present an
architecture that achieves very competitive accuracy in the ImageNet dataset (11.8\% top-5 error, 10-view test), yet
is 20\% faster than ``AlexNet'' \cite{Krizhevsky2012} (16.0\% top-5 error, 10-view test).
\end{abstract}

\section{Introduction}

Convolutional neural networks (CNNs) \cite{LeCun1989,Krizhevsky2012} have recently brought in revolutions to the computer vision area. Deep CNNs not only have been continuously advancing the image classification accuracy \cite{Krizhevsky2012,Sermanet2014,Zeiler2014,Chatfield2014,He2014,Simonyan2014,Szegedy2014}, but also play as generic feature extractors for various recognition tasks such as object detection \cite{Girshick2014,He2014}, semantic segmentation \cite{Girshick2014,Hariharan2014}, and image retrieval \cite{Krizhevsky2012,Razavian2014}.

Most of the recent advanced CNNs are more time-consuming than Krizhevsky \etal's \cite{Krizhevsky2012} original architecture in both training and testing. The increased computational cost can be attributed to the increased width\footnote{In this paper, we use ``width'' to term the number of filters in a layer. In some literatures, the term ``width'' can have different meanings.} (numbers of filters) \cite{Sermanet2014,Zeiler2014,Chatfield2014}, depth (number of layers) \cite{Simonyan2014,Szegedy2014}, smaller strides \cite{Sermanet2014,Zeiler2014,Simonyan2014}, and their combinations. Although these time-consuming models are worthwhile for advancing the state of the art, they can be unaffordable or unnecessary for practical usages. For example, an on-line commercial search engine needs to response to a request in real-time; a cloud service is required to handle thousands of user-submitted images per second; even for off-line processes like web-scale image indexing, the system needs to handle tens of billions of images in a few days. Increasing the computational power of the hardware can partially relief these problems, but will take very expensive commercial cost.
Furthermore, on smartphones or portable devices, the low computational power (CPUs or low-end GPUs) limits the speed of the real-world recognition applications. \emph{So in industrial and commercial scenarios, engineers and developers are often faced with the requirement of constrained time budget}.

Besides the test-time demands, the off-line training procedure can also be constrained by affordable time cost. The recent models \cite{Chatfield2014,He2014,Simonyan2014,Szegedy2014} take a high-end GPU or multiple GPUs/clusters one week or several weeks to train, which can sometimes be too demanding for the rapidly changing industry. Moreover, even if the purpose is purely for pushing the limits of accuracy (like for the ImageNet competition \cite{Russakovsky2014}), the maximum tolerable training time is still a major bottleneck for experimental research. While the time budget can be loose in this case, it is worthwhile to understand which factors can gain more improvement.

This paper investigates the accuracy of CNN architectures at constrained time cost during both training and testing stages.
Our investigations involve the depth, width, filter sizes, and strides of the architectures. Because the time cost is constrained, the differences among the architectures must be exhibited as trade-offs between those factors. For example, if the depth is increased, the width and/or filter sizes need to be properly reduced.
In the core of our designs is ``layer replacement'' - a few layers are replaced with some other layers that preserve time cost.
Based on this strategy, we progressively modify a model and investigate the accuracy through a series of controlled experiments. This not only results in a more accurate model with the same time cost as a baseline model, but also facilitates the understandings of the impacts of different factors to the accuracy.

From the controlled experiments, we draw the following empirical observations about the depth.
(1) The network depth is clearly of high priority for improving accuracy, \emph{even if the width and/or filter sizes are reduced to compensate the time cost}. This is not a straightforward observation even though the benefits of depth have been recently demonstrated \cite{Simonyan2014,Szegedy2014}, because in previous comparisons \cite{Simonyan2014} the extra layers are added without trading off other factors, and thus increase the complexity. (2) While the depth is important, the accuracy gets stagnant or even degraded if the depth is overly increased. This is observed even if width and/filter sizes are not traded off (so the time cost increases with depth).

Through the investigations, we obtain a model that achieves 11.8\% top-5 error (10-view test) on ImageNet \cite{Deng2009} and only takes 3 to 4 days training on a single GPU.
Our model is more accurate and also faster than several competitive models in recent papers \cite{Zeiler2014,Chatfield2014,He2014}.
Our model has 40\% less complexity than ``\emph{AlexNet}'' \cite{Krizhevsky2012} and 20\% faster actual GPU speed, while has 4.2\% lower top-5 error.

\section{Related Work}

Recently there has been increasing attention on accelerating the test-time speed of CNNs \cite{Mamalet2012,Denton2014,Jaderberg2014}. These methods approximate and simplify the trained networks, with some degradation on accuracy. These methods do not address the training time because they are all post-processes of the trained networks. Besides, when the testing time budget is given by demand, it is still desirable to find the pre-trained model subject to certain time constraints, because the speedup ratio of these acceleration methods is limited. These methods should also be applicable in our models, further speeding up the testing stage.

Constraining the network complexity is a way of understanding the impacts of the factors in the network designs. In \cite{Eigen2013}, the accuracy of tied/untied CNNs is evaluated with various width, depth, and numbers of parameters. The tied (recursive) network is designed for strictly fixing the number of parameters. In contrast, our paper investigates the accuracy while fixing the time complexity. Our paper focuses on the untied (non-recursive) CNNs trained on ImageNet, which can be more useful as generic feature extractors. We also investigate factors like filter sizes and strides.

Most recently, Szegedy \etal \cite{Szegedy2014} propose the ``inception'' layer. Their model achieves the best accuracy in ILSVRC 2014 \cite{Russakovsky2014} while its theoretical complexity is merely 50\% more than AlexNet. The inception layer is a ``multi-path'' design that concatenates several convolutional layers of various filter sizes/numbers. The principal of choosing these hyper-parameters requires further investigation, and the influence of each branch remains unclear. In our paper, we only consider ``single-path'' designs with no parallel convolutional layers, which are already faced with abundance of choices. 

\newcolumntype{x}[1]{>{\centering}p{#1pt}}
\setlength{\tabcolsep}{2pt}
\begin{table*}[t]
\begin{flushleft}
\begin{tabular}{x{10}||x{22}x{22}||x{8}|x{36}|x{16}|x{96}|x{16}|x{96}|x{16}|x{84}|c}
\hline
 ~ & top-1 & top-5 & $d$ & stage 1 & \footnotesize{pool} & stage 2 & \footnotesize{pool} & stage 3 & \footnotesize{pool} & stage 4 & \footnotesize{comp.}\\
\hline\hline
A & 37.4 & 15.9 & 5 & (7, 64)$_{/2}$ & 3$_{/3}$ & (5, 128) & 2$_{/2}$ & (3, 256)$\times$3 & & & 1\\
\hline
B & 35.7 & 14.9 & 8 & (7, 64)$_{/2}$ & 3$_{/3}$ & (5, 128) & 2$_{/2}$ & (2, 256)$\times$6 & & & 0.96\\
C & 35.0 & 14.3 & 6 & (7, 64)$_{/2}$ & 3$_{/3}$ & (3, 128)$\times$2 & 2$_{/2}$ & (3, 256)$\times$3 & & & 1.02\\
D & 34.5 & 13.9 & 9 & (7, 64)$_{/2}$ & 3$_{/3}$ & (3, 128)$\times$2 & 2$_{/2}$ & (2, 256)$\times$6 & & & 0.98\\
E & \ul{33.8} & \ul{13.3} & 11 & (7, 64)$_{/2}$ & 3$_{/3}$ & (2, 128)$\times$4 & 2$_{/2}$ & (2, 256)$\times$6 & & & 0.99\\
\hline
F & 35.5 & 14.8 & 8 & (7, 64)$_{/2}$ & 3$_{/3}$ & (5, 128) & 2$_{/2}$ & (3, 160)$\times$5+(3, 256) & & & 1\\
G & 35.5 & 14.7 & 11 & (7, 64)$_{/2}$ & 3$_{/3}$ & (5, 128) & 2$_{/2}$ & (3, 128)$\times$8+(3, 256) & & & 1\\
H & 34.7 & 14.0 & 8 & (7, 64)$_{/2}$ & 3$_{/3}$ & (3, 64)$\times$3+(3, 128) & 2$_{/2}$ & (3, 256)$\times$3 & & & 0.97\\
I & \ul{33.9} & \ul{13.5} & 11 & (7, 64)$_{/2}$ & 3$_{/3}$ & (3, 64)$\times$3+(3, 128) & 2$_{/2}$ & (2, 256)$\times$6 & & & 0.93\\
\hline
J & \textbf{32.9} & \textbf{12.5} & 11 & (7, 64)$_{/2}$ & 3$_{/3}$ & (2, 128)$\times$4 & 2$_{/2}$ & (2, 256)$\times$4 & 3$_{/3}$& {(2, 2304)+(2, 256)}& 0.98\\
\hline
\end{tabular}
\end{flushleft}
\caption{\textbf{Configurations of the models under constrained time complexity}. The notation $(s, n)$ represents the filter size and the number of filters. ``$/2$'' represents stride = 2 (default 1). ``$\times$$k$'' means the same layer configuration is applied $k$ times (not sharing weights). ``+'' means another layer is followed. The numbers in the pooling layer represent the filter size and also the stride. All convolutional layers are with ReLU.
The feature map size of stage 2 is dominantly 36$\times$36, of stage 3 is 18$\times$18, and of stage 4 (if any) is 6$\times$6. The top-1/top-5 errors (at 75 epochs) are on the validation set. The ``comp.'' is the theoretical time complexity (relative to A) computed using Eqn.(\ref{eq:time}).}
\label{tab:models}
\end{table*}

\setlength{\tabcolsep}{2pt}
\begin{table*}[t]
\begin{flushleft}
\begin{tabular}{x{10}||x{22}x{22}||x{8}|x{36}|x{16}|x{96}|x{16}|x{96}|x{16}|x{84}|c}
\hline
 & top-1 & top-5 & $d$ & stage 1 & \footnotesize{pool} & stage 2 & \footnotesize{pool} & stage 3 & \footnotesize{pool} & stage 4 & \footnotesize{comp.}\\
\hline\hline
B\str & 34.6 & 13.9 & 8 & (7, 64)$_{/2}$ & 3$_{/1}$ & (5, 128)$_{/3}$ & 2$_{/1}$ & (2, 256)$_{/2}$+(2, 256)$\times$5 & & & 0.96\\
D\str & 33.8 & 13.5 & 9 & (7, 64)$_{/2}$ & 3$_{/1}$ & (3, 128)$_{/3}$+(3, 128) & 2$_{/1}$ & (2, 256)$_{/2}$+(2, 256)$\times$5 & & & 0.98\\
E\str & 33.4 & 13.0 & 11 & (7, 64)$_{/2}$ & 3$_{/1}$ & (2, 128)$_{/3}$+(2, 128)$\times$3 & 2$_{/1}$ & (2, 256)$_{/2}$+(2, 256)$\times$5 & & & 0.99\\
J\str & \textbf{32.2} & \textbf{12.0} & 11 & (7, 64)$_{/2}$ & 3$_{/1}$ & (2, 128)$_{/3}$+(2, 128)$\times$3 & 2$_{/1}$ & (2, 256)$_{/2}$+(2, 256)$\times$3 & 3$_{/1}$ & (2, 2304)$_{/3}$+(2, 256) & 0.98\\
\hline
\end{tabular}
\end{flushleft}
\caption{\textbf{Configurations of the models with delayed subsampling of pooling layers}. The strides in the pooling layers are set as 1, and the original strides are moved to the subsequent convolutional layer.}
\label{tab:models_str}
\end{table*}

\section{Prerequisites}

\subsection{A Baseline Model}

Our investigation starts from an eight-layer model similar to an Overfeat model \cite{Sermanet2014} that is also used in \cite{Chatfield2014,He2014}. It has five convolutional (\emph{conv}) layers and three fully-connected (\emph{fc}) layers.
The input is a 224$\times$224 color image with mean subtracted. The first convolutional layer has 64 7$\times$7 filters with a stride 2, followed by a 3$\times$3 max pooling layer with a stride 3. The second convolutional layer has 128 5$\times$5 filters, followed by a 2$\times$2 max pooling layer with a stride 2. The next three convolutional layers all have 256 3$\times$3 filters. A spatial pyramid pooling (SPP) layer \cite{He2014} is used after the last convolutional layer. The last three layers are two 4096-d fc layers and a 1000-d fc layer, with softmax as the output. All the convolutional/fc layers (except the last fc) are with the Rectified Linear Units (ReLU) \cite{Nair2010,Krizhevsky2012}.
We do not apply local normalization. The details are in Table~\ref{tab:models} (A). This model is ``narrower'' (with fewer numbers of filters) than most previous models \cite{Krizhevsky2012,Howard2013,Sermanet2014,Zeiler2014,He2014}.

We train the model on the 1000-category ImageNet 2012 training set \cite{Deng2009,Russakovsky2014}.
The details of training/testing are in Sec.~\ref{sec:impl}, which mostly follow the standards in \cite{Krizhevsky2012}.
We train this model for 75 epochs, which take about 3 days. The top-1/top-5 error is 37.4/15.9 using the 10-view test \cite{Krizhevsky2012}.

In the following we will design new models with the same time complexity as this model.
We start from this model due to a few reasons. Firstly, this model mostly follows the popular ``3-stage'' designs as in \cite{Krizhevsky2012,Howard2013,Sermanet2014,Zeiler2014,He2014} - the first stage is a single convolutional layer having a few number of large-size filters (7$\times$7 or 11$\times$11) with pooling, the second stage is a single convolutional layer with 5$\times$5 filters with pooling, and the third stage is a cascade of 3$\times$3 convolutional layers. So we expect that the observations in this paper will apply for other similar models. Secondly, our baseline model has fewer filters than most previous models, so it is faster for training/testing.

Nevertheless, even though in this paper we are based on this model, some of the following observations should remain mostly valid, because the observations are drawn from several variants of the models.

\subsection{Time Complexity of Convolutions}

The total time complexity of all convolutional layers is:
\begin{equation}\label{eq:time}
O\left(\sum_{l=1}^{d}n_{l-1} \tms s_{l}^2 \tms n_{l} \tms m_{l}^2\right)
\end{equation}
Here $l$ is the index of a convolutional layer, and $d$ is the depth (number of convolutional layers). $n_{l}$ is the number of filters (also known as ``width'') in the $l$-th layer. $n_{l-1}$ is also known as the number of input channels of the $l$-th layer.
$s_{l}$ is the spatial size (length) of the filter. $m_{l}$ is the spatial size of the output feature map.
Note this time complexity applies to both training and testing time, though with a different scale. The training time per image is roughly three times of the testing time per image (one for forward propagation and two for backward propagation).

The time cost of fc layers and pooling layers is not involved in the above formulation. These layers often take 5-10\% computational time. Without further complicating the investigation, we fix the input/output dimensions of the fc and pooling layers in all the models. We only consider the trade-offs among the convolutional layers.

The theoretical time complexity in Eqn.(\ref{eq:time}), rather than the actual running time, will be the base of our network designs, because the actual running time can be sensitive to implementations and hardware. Even so, most of our models in this paper have actual running time that scales nicely with the theoretical complexity.

\section{Model Designs by Layer Replacement}

Designing a model under a constrained complexity is a complicated problem, because there are several factors involved in Eqn.(\ref{eq:time}). We simplify the cases by designing a series of ``layer replacement'' - at each time a few layers are replaced by some other layers that preserve complexity, without changing the rest layers. For the design of a layer replacement, we study the trade-offs between two factors with the rest factors unchanged.
With all these put together, we progressively modify the models and investigate the changes of accuracy through controlled experiments.

Without further complicating the cases, we will mainly trade off the factors inside a ``stage'' - a ``stage'' is defined as those layers between two nearby pooling layers. We will fix the numbers of output filters of each stage, so also fix the numbers of input channels to the next stage.

\subsection{Trade-offs between Depth and Filter Sizes}

We first investigate the trade-offs between depth $d$ and filter sizes $s$. We replace a larger filter with a cascade of smaller filters. We denote a layer configuration as:
\begin{equation}
n_{l-1} \tms s_{l}^2 \tms n_{l}
\end{equation}
which is also its theoretical complexity (with the feature map size temporarily omitted). An example replacement can be written as the complexity involved:
\begin{eqnarray}
&256 \tms 3^2 \tms 256 \nnum\\
\Rightarrow&256 \tms 2^2 \tms 256+256 \tms 2^2 \tms 256.\nonumber
\end{eqnarray}
This replacement means that a 3$\times$3 layer with 256 input/output channels is replaced by two 2$\times$2 layers with 256 input/output channels. After the above replacement, the complexity involved in these layers is nearly unchanged (slightly reduces as $(2^2+2^2)/(3^2)=8/9$).
The model B in Table~\ref{tab:models} is from the model A using this replacement.

Similarly, we replace the 5$\times$5 filters by two 3$\times$3 filters. The complexity involved is:
\begin{eqnarray}
&64 \tms 5^2 \tms 128 \nnum\\
\Rightarrow&64 \tms 3^2 \tms 128+128 \tms 3^2 \tms 128.\nonumber
\end{eqnarray}
This replacement also approximately preserves the complexity (increasing by $\sim$8\% of this layer).
The models C and D in Table~\ref{tab:models} are from A and B using this replacement.
We can further replace the 5$\times$5 filters by four 2$\times$2 filters:
\begin{eqnarray}
&64 \tms 5^2 \tms 128 \nnum\\
\Rightarrow&64 \tms 2^2 \tms 128+(128 \tms 2^2 \tms 128)\times3.\nonumber
\end{eqnarray}
The model E in Table~\ref{tab:models} is from B with this replacement.

In case when 2$\times$2 filters are used, the feature map size cannot be preserved strictly.
To address this issue, we consider two sequential 2$\times$2 layers: we use no padding in the first 2$\times$2 layer, and pad one pixel (on each side) in the next 2$\times$2 layer. This reduces the feature map size by one pixel after the first layer, but restores the original feature map size after the second layer. This also slightly reduces the complexity of using 2$\times$2 layers due to the reduced feature map sizes in the first layer. Because we cannot strictly preserve time complexity, we also show the complexity (relative to A) of the models in Table~\ref{tab:models}.

\begin{figure}[t]
\begin{center}
\includegraphics[width=1.0\linewidth]{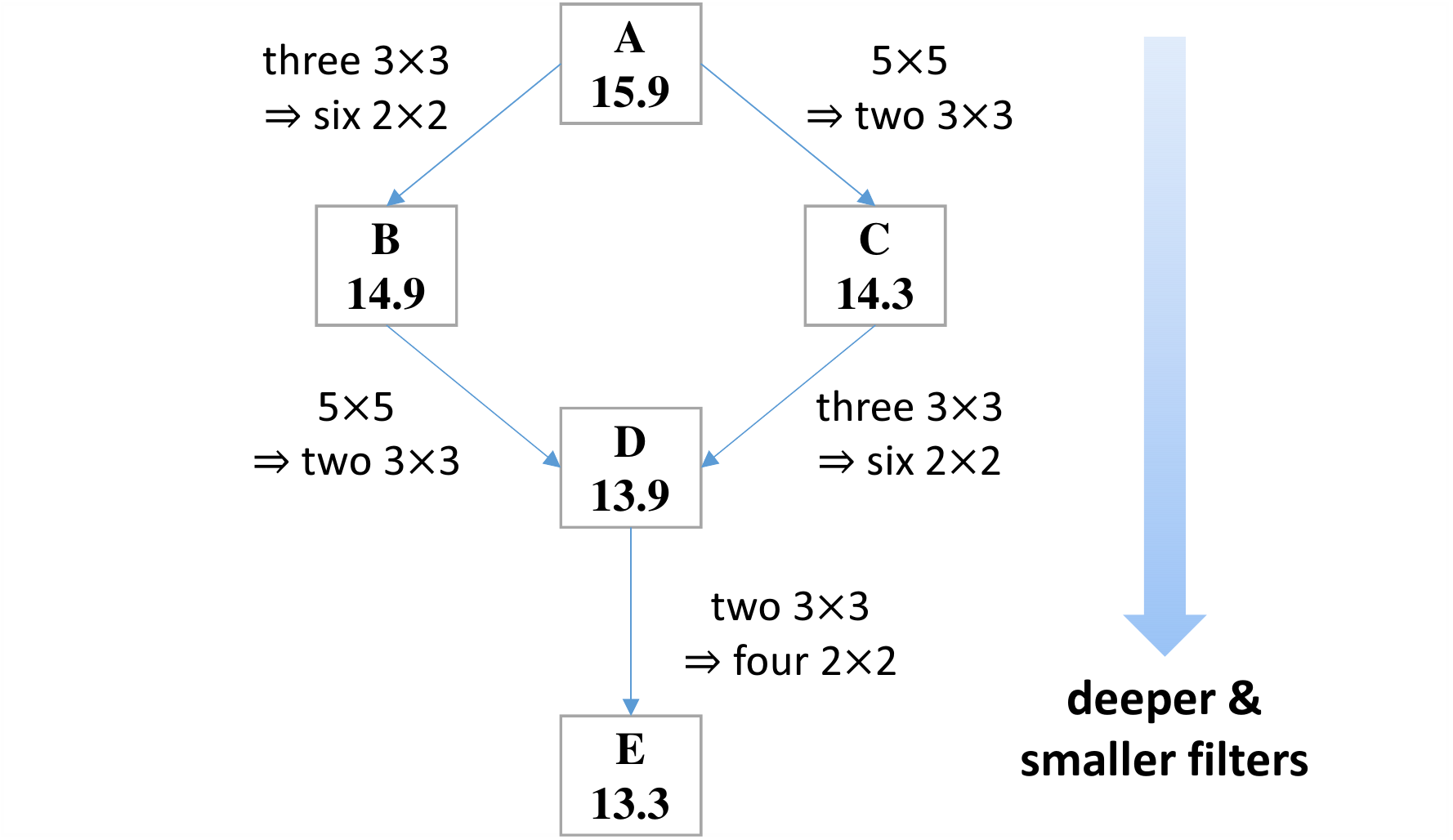}
\end{center}
\caption{The relations of the models about \textbf{depth and filter sizes}.}
\label{fig:depth_size}
\end{figure}

Fig.~\ref{fig:depth_size} summarizes the relations among the models A, B, C, D, and E.
Table~\ref{tab:models} shows their detailed structures and the top-1/top-5 errors (10-view test).
These results show that the depth is more important than the filter sizes. When the time complexity is roughly the same, the deeper networks with smaller filters show better results than the shallower networks with larger filters.

\begin{figure}[t]
\begin{center}
\includegraphics[width=1.0\linewidth]{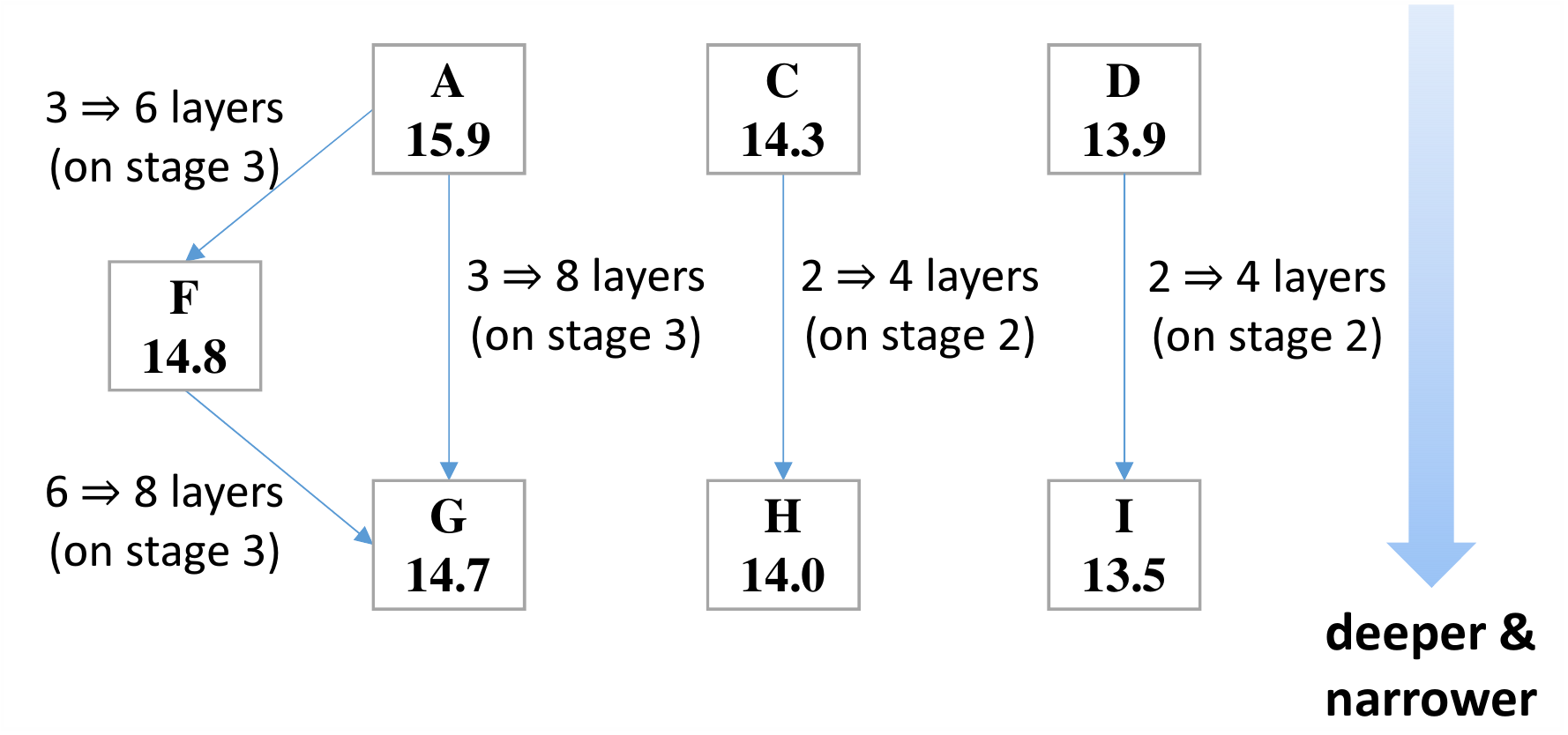}
\end{center}
\caption{The relations of the models about \textbf{depth and width}.}
\label{fig:depth_width}
\end{figure}

\begin{figure}
\begin{center}
\includegraphics[width=1.0\linewidth]{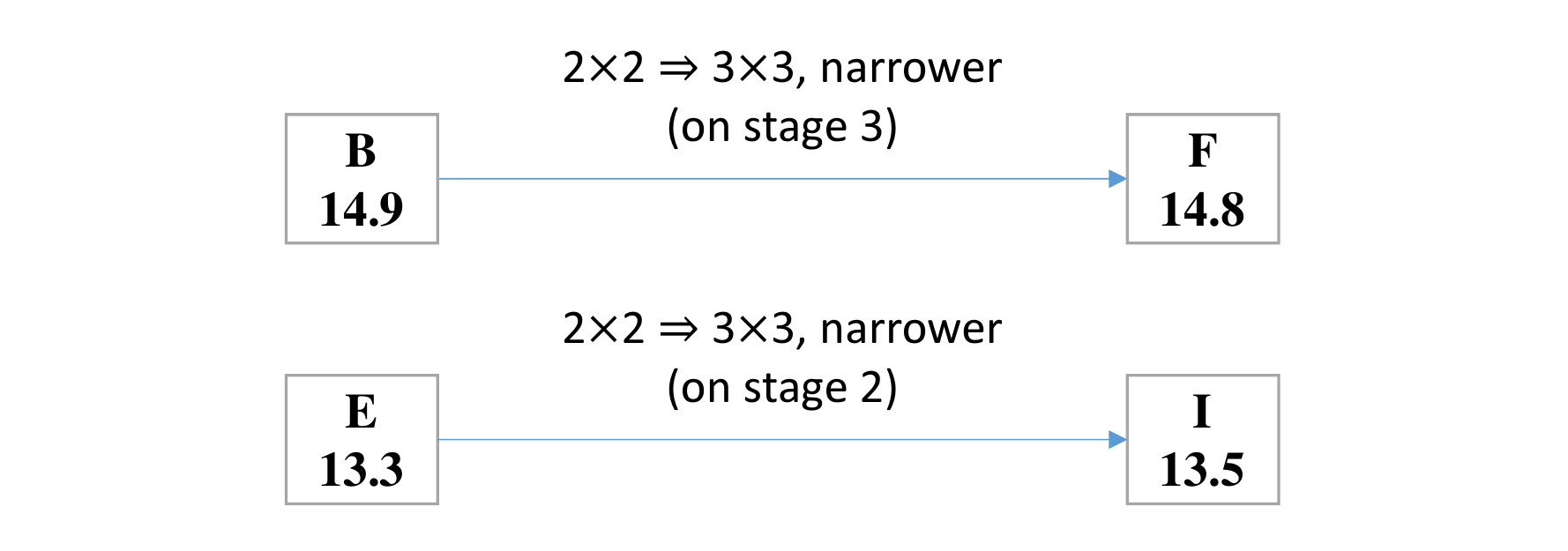}
\end{center}
\caption{The relations of the models about \textbf{width and filter sizes}.}
\label{fig:width_size}
\end{figure}

\subsection{Trade-offs between Depth and Width}

Next we investigate the trade-offs between depth $d$ and width $n$. We increase depth while properly reducing the number of filters per layer, without changing the filter sizes.
We replace the three 3$\times$3 layers in A with six 3$\times$3 layers (model F). The complexity involved is:
\begin{small}
\begin{eqnarray}
&128 \tms 3^2 \tms 256 + (256 \tms 3^2 \tms 256)\times2 \nnum\\
=&128 \tms 3^2 \tms 160+(160 \tms 3^2 \tms 160)\times4+160 \tms 3^2 \tms 256.\nonumber
\end{eqnarray}
\end{small}Here we fix the number of the input channels (128) of the first layer, and the number of output filters (256) of the last layer, so avoid impacting the previous/next stages. With this replacement, the width reduces from 256 to 160 (except the last layer). This replacement is designed such that it exactly preserves the complexity, so we use the notation ``$=$''.

We can also replace the three 3$\times$3 layers in A with nine 3$\times$3 layers (model G):
\begin{eqnarray}
&128 \tms 3^2 \tms 256 + (256 \tms 3^2 \tms 256)\times2 \nnum\\
=&(128 \tms 3^2 \tms 128)\times8+128 \tms 3^2 \tms 256.\nonumber
\end{eqnarray}
This replacement also exactly preserves the complexity.

We also consider a replacement in stage 2 on the models C and D. We replace the two 3$\times$3 layers by four layers. The complexity involved is
\begin{eqnarray}
&64 \tms 3^2 \tms 128+128 \tms 3^2 \tms 128\nnum\\
\Rightarrow&(64 \tms 3^2 \tms 64)\times3+64 \tms 3^2 \tms 128.\nonumber
\end{eqnarray}
This gives the models H and I.

Fig.~\ref{fig:depth_width} summarizes the relations among the models of A, F, G, H, and I.
Table~\ref{tab:models} shows the details and the top-1/top-5 error.
We find that increasing the depth leads to considerable gains, even the width needs to be properly reduced. The models F and G has 14.8 and 14.7 top-5 error respectively, much lower than the error of A (15.9). The models H and I also have lower error than C and D.

It is worth noticing that G can also be considered as a deeper and narrower version of F (see Fig.~\ref{fig:depth_width}). But G is only better than F marginally. The improvement due to increased depth becomes saturated.
We will further investigate a related issue in Sec.~\ref{sec:deeper}.

\subsection{Trade-offs between Width and Filter Sizes}

We can also fix the depth and investigate the trade-offs between width and filter sizes. Some of the above models can actually be considered as undergone such replacement.

The models B and F exhibit this trade-off on the last six convolutional layers:
\begin{small}
\begin{eqnarray}
&128 \tms 2^2 \tms 256+(256 \tms 2^2 \tms 256)\times5  \nnum\\
\Rightarrow&128 \tms 3^2 \tms 160+(160 \tms 3^2 \tms 160)\times4+160 \tms 3^2 \tms 256.\nonumber
\end{eqnarray}
\end{small} This means that the first five 2$\times$2 layers with 256 filters are replaced with five 3$\times$3 layers with 160 filters. The errors are comparable: 35.7/14.9 for B and 35.5/14.8 for F.

Similarly, the models E and I have this trade-off:
\begin{eqnarray}
&(64 \tms 3^2 \tms 64)\times3 + 64 \tms 3^2 \tms 128\nnum\\
\Rightarrow&64 \tms 2^2 \tms 96+96 \tms 2^2 \tms 128+(128 \tms 2^2 \tms 128)\times2\nonumber.
\end{eqnarray}
This means that the four 3$\times$3 layers with 64 filters (128 in the last) are replaced with four 2$\times$2 layers with 128 filters.
The top-1/top-5 errors are also comparable (33.8/13.3 for E and 33.9/13.5 for I).

Fig.~\ref{fig:width_size} shows the relations of these models.
Unlike the depth that has a high priority, the width and filter sizes (3$\times$3 or 2$\times$2) do not show apparent priorities to each other.

\subsection{Is Deeper Always Better?}
\label{sec:deeper}

The above results have shown the priority of depth for improving accuracy. With the above trade-offs, we can have a much deeper model if we further decrease width/filter sizes and increase depth. However, in experiments we find that the accuracy is stagnant or even reduced in some of our very deep attempts. There are two possible explanations: (1) the width/filter sizes are reduced overly and may harm the accuracy, or (2) overly increasing the depth will degrade the accuracy even if the other factors are not traded.
To understand the main reason, \emph{in this subsection we do not constrain the time complexity} but solely increase the depth without other changes.

We add several duplicates of the last convolutional layers on the model D. Each extra layer has 256 2$\times$2 filters. The other layers are not modified, so the time cost will be increased. Table~\ref{tab:depth} shows the error \vs the depth. We find that the errors not only get saturated at some point, but get worse if going deeper. We also find that the degradation is not due to over-fitting, because the training errors are also clearly worse. Similar phenomena are also observed when 3$\times$3 filters are added in the models C and G.

This experiment shows that overly increasing depth can harm the accuracy, even if the width/filter sizes are unchanged. So it is not beneficial to further increase depth with width/filter sizes decreased.

We have also tested the Network-In-Network (NIN) method \cite{Lin2013} on our models. The NIN is essentially a 1$\times$1 convolutional layer following some other layers. We add a 1$\times$1 layer (preserving the input/output dimensions) after each of the 2$\times$2 and 3$\times$3 convolutional layers in the models B and D. This operation increases the depth. We find that the top-1/5 errors of B are increased to 37.8/16.5 (from 35.7/14.9), and those of D are increased to 36.9/15.7 (from 34.5/13.9).
We believe that this is also due to the over increase of depth. We have also tried 1$\times$1 layers that reduce dimensions by 1/2, but found no better result.

\setlength{\tabcolsep}{8pt}
\begin{table}[t]
\small
\begin{center}
\begin{tabular}{c|ccccc}
\hline
model & D & D+2 & D+4 & D+6 & D+8\\
\hline
top-1 & 34.5 & 34.0 & 33.9 & 34.0 & 34.2\\
top-5 & 13.9 & 13.6 & 13.4 & 13.5 & 13.6\\
\hline
\end{tabular}
\end{center}
\caption{Error rates of models with increased depth. The model ``D+$i$'' means (2, 256)$\times i$ are added on the last stage of the model D. In this table, \emph{we do not constrain the time complexity}, so the deeper models are slower.}
\label{tab:depth}
\end{table}

\subsection{Adding a Pooling Layer}

In the above, the feature map size $m_l$ of each stage is unchanged (or nearly unchanged if there is padding). Next we study a trade-off between the feature map size and width.

Because the feature map sizes are mainly determined by the strides of all previous layers, modifying the stride of an earlier layer may require to adjust all the subsequent layers to preserve complexity.
So we consider a simple and controlled case: add a pooling layer after the stage 3 and move a few previous convolutional layers to this new stage, while most layers are kept unchanged.

The model J is obtained from the model E using this way. On the stage 3, we add a max pooling layer whose stride and filter size are 3. Two convolutional layers are moved from the stage 3 to the new stage 4. This allows to use more filters on the stage 4. The complexity involved is:
\begin{eqnarray}
&256 \tms 2^2 \tms 256+256 \tms 2^2 \tms 256 \\
=&(256 \tms 2^2 \tms 2304 + 2304 \tms 2^2 \tms 256) / 3^2\nonumber,
\end{eqnarray}
Here the factor $1/3^2$ is because of the stride of the added pooling layer.
The replacement shows that the first convolutional layer on the new stage has much more filters (2304) without changing the time complexity\footnote{Due to the implementation of 2$\times$2 filters, the involved feature maps are 17$\times$17 and 18$\times$18 before replacement, and 5$\times$5 and 6$\times$6 after replacement. So the involved complexity slightly reduces after replacement.}.

Note that we always keep the number of filters in the last convolutional layer as 256. Because the subsequent SPP layer \cite{He2014} will always generate fixed-length outputs, the fc layers have the same dimensionality as all other models.

As in Table~\ref{tab:models}, the model J results in 32.9/12.5 error rates. This is considerably better than the counterpart (E, 33.8/13.3). This gain can be attributed to the increased number of filters on a layer. Besides, the feature maps in the stage 4 have coarser resolution, and the filters can capture more non-local but still translation-invariant information.

\subsection{Delayed Subsampling of Pooling Layers}

In previous usages of max pooling \cite{Krizhevsky2012,Ciresan2012}, this layer has been playing two different roles: (i) lateral suppression (max filtering) that increases the invariance to small local translation, and (ii) reducing the spatial size of feature maps by subsampling. Usually a max pooling layer plays the two roles simultaneously (stride $>1$).

Here we separate these roles by setting stride $=1$ on the max pooling layer. To preserve the complexity of all layers that follow, we use a stride $>1$ on the subsequent convolutional layer. This stride equals to the original stride of the pooling layer. This operation does not change the complexity of all convolutional layers.

The models B\str/D\str/E\str/J\str in Table~\ref{tab:models_str} are the variants of B/D/E/J with this modification.
Table~\ref{tab:delayed} summarizes the results.
Consistent gains are observed. For the models of J and J\str, this operation reduces the top-1/5 error from 32.9/12.5 to 32.2/12.0.

\setlength{\tabcolsep}{8pt}
\begin{table}[h]
\small
\begin{center}
\begin{tabular}{c|cccc}
\hline
 delayed & B & D & E & J \\
\hline
no & 14.9 & 13.9 & 13.3 & 12.5\\
\textbf{yes} & \ul{13.9} & \ul{13.5} & \ul{13.0} & \ul{\textbf{12.0}}\\
\hline
\end{tabular}
\end{center}
\caption{Top-5 error rates with/without delayed subsampling of the pooling layers. The numbers are from Table~\ref{tab:models} and Table~\ref{tab:models_str}.}
\label{tab:delayed}
\end{table}

\setlength{\tabcolsep}{3pt}
\begin{table*}[t]
\begin{center}
\begin{tabular}{x{96}||x{36}|x{36}||x{48}|c}
\hline
 model & top-1 & top-5 & complexity & seconds/mini-batch \\
\hline
AlexNet \cite{Krizhevsky2012}, our impl. & 37.6 & 16.0 & 1.4$\times$ & 0.50 (1.2$\times$)\\
ZF (fast) \cite{Zeiler2014}, our impl. & 36.0 & 14.8 & 1.5$\times$ & 0.54 (1.3$\times$)\\
CNN-F \cite{Chatfield2014} & - & 16.7 & 0.9$\times$ & 0.30 (0.7$\times$)\\
SPPnet (ZF5) \cite{He2014} & 35.0 & 14.1 & 1.5$\times$ & 0.55 (1.3$\times$)\\
\hline
ours & \textbf{31.8} & \textbf{11.8} & 1 & \textbf{0.41}\\
\hline
\end{tabular}
\end{center}
\caption{Comparisons with \textbf{fast models} in ImageNet. All results are based on the \textbf{10-view testing} of a single model. The ``complexity'' column is the time complexity of convolutional layers as computed in (\ref{eq:time}), shown by relative numbers compared with the model J\str. The running time is the actual training time per 128 views, with the relative numbers shown in the brackets. The actual testing time is about 1/3 of the actual training time.}
\label{tab:imagenet_fast}
\end{table*}

\setlength{\tabcolsep}{3pt}
\begin{table*}[t]
\begin{center}
\begin{tabular}{x{96}||x{36}|x{36}||x{48}|c}
\hline
 model & top-1 & top-5 & complexity & seconds/mini-batch \\
\hline
CNN-M \cite{Chatfield2014} & - & 13.7 & 2.1$\times$ & 0.71 (1.7$\times$)\\
CNN-S \cite{Chatfield2014} & - & 13.1 & 3.8$\times$ & 1.23 (3.0$\times$)\\
SPPnet (O5) \cite{He2014} & 32.9 & 12.8 & 3.8$\times$ & 1.24 (3.0$\times$)\\
SPPnet (O7) \cite{He2014} & 30.4 & 11.1 & 5.8$\times$ & 1.85 (4.5$\times$)\\
\hline
ours & 31.8 & 11.8 & 1 & \textbf{0.41}\\

\hline\multicolumn{5}{c}{}\\[-1.5ex]
\multicolumn{5}{c}{+ data augmentation for training}\\
\hline
VGG-16 \cite{Simonyan2014}$^\dagger$ & 28.1 & 9.3 & 20.1$\times$ & 9.6 (23.4$\times$)\\
GoogLeNet \cite{Szegedy2014} & - & 9.2 & 2.1$\times$ & 3.2 (7.8$\times$)\\
\hline
\end{tabular}
\end{center}
\caption{Comparisons with \textbf{accurate models} in ImageNet. All results are based on the \textbf{10-view testing} of a single model. The running time is the actual training time per 128 views.
$^\dagger$The 10-view results of VGG-16 is computed using the model released by the authors.}
\label{tab:imagenet_accurate}
\end{table*}

\vspace{-6pt}
\subsection{Summary}

From the baseline A (37.4/15.9), we arrive at the model J\str whose top-1/5 errors are \textbf{32.2} and \textbf{12.0}. The top-1 error is reduced by \textbf{5.2}, and top-5 by \textbf{3.9}.
The convolution complexity is not increased.
Actually, the model J\str has a slightly smaller complexity than A, mainly due to the usage of 2$\times$2 filters. Also note that the input/output dimensions of the fc layers are exactly the same for all models.

\section{Implementation Details}
\label{sec:impl}

We implement the models using the \emph{cuda-convnet2} library\footnote{\url{https://code.google.com/p/cuda-convnet2/}} \cite{Krizhevsky2014} with our modifications. All experiments are run on a single GPU.
All the models are trained/tested based on the same implementation as follows.

\vspace{6pt}
\noindent\textbf{Data Augmentation.} During training, the 224$\times$224 image is randomly cropped from a full image whose shorter side is 256, as in \cite{Howard2013}. A half of random samples are flipped horizontally. The color altering \cite{Krizhevsky2012} is also used. No other data augmentation, such as scale jittering \cite{Howard2013,Simonyan2014,He2014}, is used.
During testing, the 10 views are those cropped from the center or the four corners, with their flipped versions.

\vspace{6pt}
\noindent\textbf{Settings.}
The mini-batch size is 128.
The learning rate is 0.01 for 10 epochs, 0.001 for 60 epochs, and 0.0001 for the rest. The weight decay is 0.0005, and momentum 0.9. The weights are randomly initialized from zero-mean Gaussian distributions \cite{Glorot2010}. The biases are all initialized as zero. Dropout (50\%) is used in the first two fc layers.

We pad no pixel for the 7$\times$7 filters, and pad 2 pixels for the 5$\times$5 filters and 1 for the 3$\times$3 filters. For two sequential 2$\times$2 filters, we do not pad the first and pad 1 pixel on the second. For some layers before/after a subsampling layer (stride $>$ 1), we need to adjust the padding such that: the feature map sizes on the stage 2 is dominantly 36$\times$36, on the stage 3 18$\times$18, and on the stage 4 (if any) 6$\times$6.

Between the last convolutional layer and the subsequent fc layer, we use the spatial pyramid pooling (SPP) as in \cite{He2014}. The pyramid has 4 levels: \{6$\times$6, 3$\times$3,  2$\times$2, 1$\times$1\}, totally 50 bins.
Each level is implemented as a sliding-window pooling layer with a proper filter size and a stride \cite{He2014}.
The four levels are concatenated and fed into the fc layer.

\section{Comparisons}
\label{sec:comp}

We compare our models with the existing models on the 1000-category ImageNet 2012 dataset \cite{Deng2009,Russakovsky2014}. For fair and controlled comparisons, we focus on the models that are trained/tested using the same way of data augmentation. Advanced data augmentation is critical to the accuracy \cite{Howard2013,He2014,Simonyan2014,Szegedy2014}, but makes it difficult to compare the architectures alone. In this paper, we adopt the same data augmentation as Chatfield \etal's paper \cite{Chatfield2014} and He \etal's single-size version (no scale-augmentation) \cite{He2014}, so we cite numbers from these two papers. For some earlier papers including Krizhevsky \etal's \cite{Krizhevsky2012} and Zeiler and Fergus's \cite{Zeiler2014}, we re-implement their models using the same setting as ours. In our re-implementations of \cite{Krizhevsky2012,Zeiler2014}, \emph{the accuracy is better than those reported in these papers}, mainly because the views are randomly sampled from the full image, rather than from the center 256$\times$256 part \cite{Howard2013}.

Following previous papers \cite{Krizhevsky2012,He2014}, in the comparisons below we train the models for 90 epochs (15 more epochs than in Table~\ref{tab:models} and~\ref{tab:models_str}). The model J\str has a reduced top-1 error as \textbf{31.8\%}, and top-5 as \textbf{11.8\%}.

We also evaluate the actual running time, which is the training time per mini-batch (128) on a single Nvidia Titan GPU (6GB memory). The testing time is about $1/3$ of the training time.
All layers (including fc, pooling, and others) are involved in the actual running time.
Our model takes 0.41 second training per mini-batch, or totally 3 to 4 days training. The testing time is 1.0 millisecond per view.

\vspace{6pt}
\noindent\textbf{Comparisons with Fast Models.}
In Table~\ref{tab:imagenet_fast} we compare with the ``fast'' models in the literature. Here ``AlexNet'' is our re-implementation of \cite{Krizhevsky2012}, except that we ignore the two-GPU splitting; ZF (fast) is our re-implementation of the smaller model in Zeiler and Fergus's \cite{Zeiler2014}; {CNN-F} is the fast model in Chatfield \etal's \cite{Chatfield2014}, which has a structure similar to AlexNet but has fewer filters; SPPnet (ZF5) \cite{He2014} is the ZF (fast) model combined with the SPP layer.
Table~\ref{tab:imagenet_fast} also lists the convolutional time complexity.

Table~\ref{tab:imagenet_fast} shows that our model is more accurate than these fast models by substantial margins. Our model has a top-5 error rate 4.2\% lower than AlexNet and a top-1 error rate 5.8\% lower, but has 40\% less complexity. The actual running time of our model is 20\% faster than AlexNet. The difference between the (convolutional) complexity and the actual running time is mainly due to the overhead of the fc and pooling layers.

\vspace{6pt}
\noindent\textbf{Comparisons with Accurate Models.}
In Table~\ref{tab:imagenet_accurate} (top) we compare with the state-of-the-art models in terms of accuracy. Here CNN-M \cite{Chatfield2014} is a model similar to ZF (fast) but with more filters; CNN-S \cite{Chatfield2014} is a model similar to the Overfeat structure \cite{Sermanet2014}; SPPnet (O5/7) \cite{He2014} are models similar to the Overfeat structure \cite{Sermanet2014} but combined with the SPP layer.
The reported results of these models are cited from the papers \cite{Chatfield2014,He2014} using the same augmentation as ours.

Table~\ref{tab:imagenet_accurate} (top) shows that our model is more accurate than CNN-M, CNN-S, and SPPnet (O5), and is also faster. SPPnet (O7) is 0.7\% better than ours in top-5 error, but requires 4.5$\times$ running time.

In Table~\ref{tab:imagenet_accurate} (bottom) we further compare with the winners of ImageNet Large Scale Recognition Challenge (ILSVRC) 2014 \cite{Russakovsky2014}. We compare on the 10-view testing results of a single model. These competitors were trained using more sophisticated data-augmentation, including different implementations of scale augmentation \cite{Simonyan2014,Szegedy2014}. The extra data augmentation should partially contribute to the gaps.

Here VGG-16 is the 16-layer model in \cite{Simonyan2014}. Because the 10-view testing result is not reported in \cite{Simonyan2014}, we download their released model\footnote{\url{www.robots.ox.ac.uk/~vgg/research/very_deep/}} and evaluate the 10-view results. The 10 views used here are 224$\times$224 cropped from resized images whose shorter side is 384 (if this number is 256 as in other cases, the accuracy is poorer). This model shows compelling results (28.1/9.3), but has over 20$\times$ larger complexity than our model. To evaluate its actual running time, we use the Caffe library \cite{Jia2014} and one K40 GPU with 12 GB memory. The cuda-convnet2 library does not support a model of this large scale.
We set the mini-batch size as 64 to fit the model into memory, so the time reported is of two mini-batches. The actual running time is 9.6s, 23$\times$ of ours.

The GoogLeNet model \cite{Szegedy2014} is the winner of ILSVRC 2014.
It shows outstanding accuracy with top-5 error 9.15\% \cite{Szegedy2014}. It also has a nice convolution complexity (2.1$\times$ of ours). But the inception structure is less friendly to the current GPU implementations. We evaluate the actual running time in the K40 GPU using the Caffe library. The running time is 3.2 seconds per mini-batch, which is 7.8$\times$ of ours. Nevertheless, this model is very CPU-friendly.

\section{Conclusion and Future Work}

Constrained time cost is a practical issue in industrial and commercial requirements. Fixing the time complexity also helps to understand the impacts of the factors. We have evaluated a series of models under constrained time cost, and proposed models that are fast for practical applications yet are more accurate than existing fast models.

Another practical issue is about memory. Our deeper models consume more memory than our baseline model. Fortunately, the memory limit is less demanding for test-time evaluation, because there is no backward propagation. The memory issue can also be partially addressed by using a smaller mini-batch size.
Nevertheless, it is still an interesting topic of investigating CNNs at constrained memory cost in the future.

{\small
\bibliographystyle{ieee}
\bibliography{cnn_cost}
}

\end{document}